\pdfoutput=1

\documentclass[11pt]{article}

\usepackage[]{ACL2023}

\usepackage{adjustbox}
\usepackage{makecell}
\usepackage{multicol}
\usepackage{multirow}
\usepackage{hhline}
\usepackage{booktabs}
\usepackage{color}
\usepackage{xcolor}
\usepackage{amssymb}
\usepackage{subfigure} 
\usepackage{amsmath}
\usepackage{times}
\usepackage{latexsym}
\usepackage{interval}
\usepackage{array}
\usepackage{colortbl}
\usepackage{xcolor}

\usepackage[T1]{fontenc}

\usepackage[utf8]{inputenc}

\usepackage{microtype}

\usepackage{inconsolata}
\usepackage{cleveref}
\usepackage{makecell}
\usepackage{booktabs}

\crefname{section}{§}{§§}

\usepackage{multirow}
\usepackage{arydshln}

\newcommand{\taskname}{Zero-Shot Instruction Following}

\newcommand{\tkinstruct}{\textsc{TK-Instruct}}
\newcommand{\methodname}{\textsc{Pick\&Rank}}
\newcommand{\exactmatch}{\textsc{ExactMatch}}
\newcommand{\rougeL}{\textsc{ROUGE-L}}
\newcommand{\dataname}{\textsc{Super-NaturalInstru}}

\definecolor{MyBlue}{HTML}{4791d9}
\definecolor{MyOrange}{HTML}{ea7827}

\newcommand{\PreserveBackslash}[1]{\let\temp=\\#1\let\\=\temp}
\newcolumntype{C}[1]{>{\PreserveBackslash\centering}p{#1}}
\newcolumntype{R}[1]{>{\PreserveBackslash\raggedleft}p{#1}}
\newcolumntype{L}[1]{>{\PreserveBackslash\raggedright}p{#1}}

\title{Toward Zero-Shot Instruction Following}

\author{
  Renze Lou \quad
  \and 
  Wenpeng Yin
  \\
  The Pennsylvania State University
  \\
  {\small \texttt{\{renze.lou, wenpeng\}@psu.edu}}
}

\begin{document}
\maketitle

\begin{abstract}

This work proposes a challenging yet more realistic setting for zero-shot cross-task generalization: \emph{zero-shot instruction following}, presuming the existence of a paragraph-style task definition while no demonstrations exist. To better learn the task supervision from the definition, we propose two strategies: first, to automatically find out the critical sentences in the definition; second, a ranking objective to force the model to generate the gold outputs with higher probabilities when those critical parts are highlighted in the definition. The joint efforts of the two strategies yield state-of-the-art performance on the \dataname~\cite{wang2022benchmarking}.\footnote{Code: \url{https://github.com/RenzeLou/Pick-Rank}}


\end{abstract}

\section{Introduction}
\label{sec:intro}

With the rapid evolutions of the pre-training techniques, large language models (LLMs), such as GPT-3~\cite{brown2020language} and ChatGPT~\cite{openai2022chatgpt}, are found to be capable of handling various novel NLP tasks by following in-context instructions~\cite{radford2019language}.\footnote{Task instructions can be any textual expressions, e.g., task names, short sentences, or paragraphs, that describe the task semantics; prompts are the special case of instructions~\cite{lou2023prompt}.}
Typically, a formal task instruction consists of two components: (1) a task definition that describes the task intent; (2) a few labeled examples to demonstrate this intent (i.e., demonstrations). Then the problem is often named as ``\emph{\(k\)-shot instruction following}'', where \(k\) is the example size. Due to the performance superiority of the in-context examples~\cite{lampinen2022can,gu2022robustness}, prior research has predominantly relied on demonstrations, allocating relatively limited attention toward effectively utilizing task definitions; we refer to this setting as ``demonstration-driven instruction following''~\cite{min2022noisy,min-etal-2022-metaicl,hu2022context}.

Notwithstanding the surprising results, this phenomenon could manifest as an instance of overestimated progress. Two reasons: firstly, \textbf{demonstrations are usually hard to be crafted in real-world applications}. Since LLMs are becoming helpful daily-task assistants and most end-users are non-experts~\cite{vicuna2023,xie2023adaptive,xie2024Travel}, it is usually exhausting and unrealistic for users to design \textit{concrete} demonstrations for every daily task, especially for those tasks that require specific domain knowledge. Secondly, as \citet{gu2022robustness} concluded, so far, \textbf{we still lack a method to effectively learn from instructions to solve tasks without demonstrations} for various reasons. For example, \citet{khashabi2022reframing} showed that the models constantly ignored the crucial information emphasized in the definition (e.g., an output constraint that asks models to ``\textit{generate no more than five words}''); \citet{webson-pavlick-2022-prompt} found that the models always struggled to truly understand the content of the definition.

To more effectively utilize the task definition, this work studies a more challenging setting: \emph{zero-shot instruction following}.  Technically, our approach consists of two strategies.\footnote{In the rest of the paper, we use the terms ``definition'' and ``instruction'' alternately, when examples are unavailable.} (i) Strategy I: automatically learn the critical task-relevant information from the lengthy definition to help the model better grasp the instruction. (ii) Strategy II: to make the model truly distinguish instructions that are specified by the critical information or not, we set a ranking-based training objective. Given instructions with critical information highlighted, this ranking strategy forces the model to generate ground-truth outputs with higher probabilities than instructions otherwise. Our system, \methodname, achieves state-of-the-art on the benchmark, \dataname~\cite{wang2022benchmarking}.
\begin{figure*}[!ht]
	\begin{center}
		\centering
		\includegraphics[width=0.85\linewidth, trim=10 10 0 0]{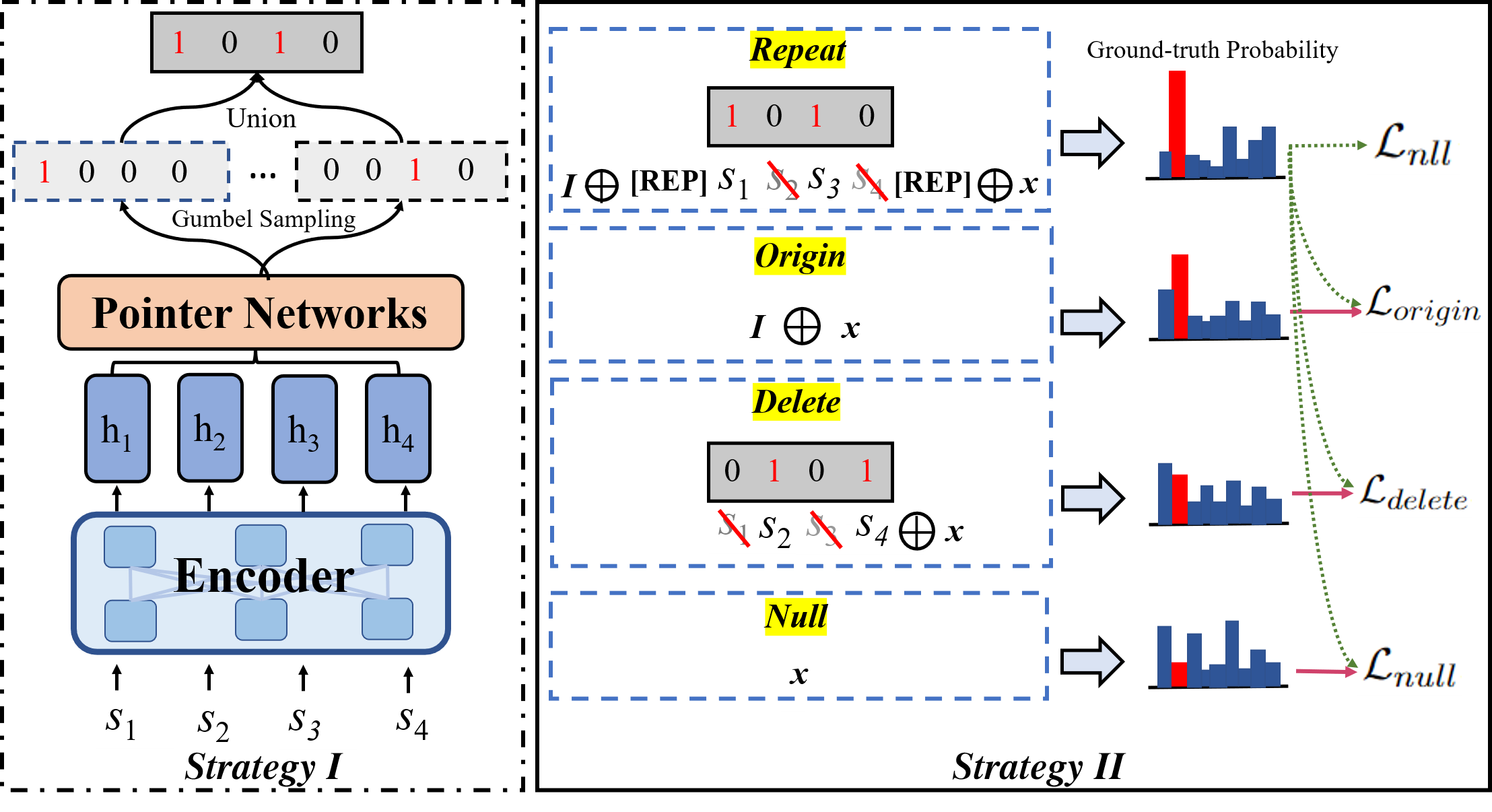}
	\end{center}
	\caption{The illustration of our \methodname. Two main components: \textbf{Strategy I} (\textsc{Pick}) and \textbf{Strategy II} (\textsc{Rank}). Strategy I aims to predict a binary value for each sentence in a definition, indicating whether a sentence is crucial. The outputs of Strategy I are used to construct instructions with different sufficiencies, e.g., ``\textit{Repeat}'' represents the most beneficial instructions where the crucial sentences are repeated. Strategy II then drives the LMs to generate higher ground-truth probabilities on the more beneficial instructions. The whole system is optimized end-to-end.}
	\label{fig:overall architecture}
\end{figure*}

\section{Related Work}
\label{sec:relatedwork}

\paragraph{Prompt \& In-context Learning.}

Prompting techniques usually acquire answers from large language models (LLMs) after rewriting the original task input into a LLM-oriented format. Impressive progress has been observed in various NLP tasks, such as question answering \cite{radford2019language}, text generation \cite{schick2021few},  information extraction \cite{wang2022instructionner,sun2024umie}, etc. \citet{brown2020language} further developed in-context learning (ICL): GPT-3 achieved competitive few-shot results without parameter tuning by prepending a prompt with a few demonstrations to new inputs.  Follow-up work delved into improving ICL, including how to choose better demonstrations \cite{rubin2021learning,lu2022fantastically}, how to formulate the tasks \cite{zhao2021calibrate,min2022noisy}, etc. However, the short and simplistic nature of the prompts makes it difficult to express NLP tasks of diverse complexities~\cite{chen2022knowprompt}. Our work tries to learn from instructions that describe the task semantics in more detail, such as Amazon MTurk instructions.

\paragraph{Follow Human-annotation Instructions.} Prompts are more friendly for LLM to emit outputs. In the real world, humans describe tasks using paragraph-style instructions, such as crowd-sourcing guidelines. This type of instruction has recently attracted much attention, including increasingly larger datasets \cite{mishra2022cross, wang2022benchmarking}, new learning problems~\cite{yin2022contintin} and applications~\cite{Zhang2023MagicBrush}, etc. To achieve cross-task generalization given instructions, prior systems trained a text-to-text model on a long sequence of text, i.e., concatenating task definition, demonstrations, and all other resources~\cite{lou2023prompt}. We ignore demonstrations and focus on the supervision extraction from task definitions.

\section{Problem Definition \& Our Approach}
\label{sec:method}

We study zero-shot instruction following in a cross-task generalization setting, where evaluation tasks are unseen in training.

\paragraph{\taskname:} Three task sets: \textsc{training tasks}, \textsc{dev tasks}, and \textsc{test tasks}. There are no overlapping tasks among them.  Each task $T$ has its instruction $I$ and a collection of labeled examples $D$ =\{($x$, $y$)\}. $x$: input; $y$: gold output of $x$ under  $I$. $I$ is a short paragraph consisting of $n$ sentences, i.e., $I=\{s_1, s_2, \cdots, s_n\}$. No examples exist in $I$. $D$ of \textsc{dev tasks} and \textsc{test tasks} are only used for evaluation.
As shown in Figure~\ref{fig:overall architecture}, we adopt two strategies to better leverage the supervision in $I$.

\paragraph{Strategy I: picking critical sentences of instructions.}
Given the instruction $I=\{s_1, \cdots, s_n\}$, the goal of this phase is to learn a binary value for each $s_i$, indicating that if $s_i$ is critical for the task $T$. We expect to select $k$ most critical sentences.

As shown in Figure~\ref{fig:overall architecture}, we train a Pointer Network~\cite{vinyals2015pointer} to select critical sentences from the input automatically. First, we concatenate all \{$s_i$\} in $I$ as the encoder input to learn a hidden vector $h_i$ for each $s_i$ as: $h_i = \textrm{Encoder}(s_i|I)$, where $h_i\in\mathcal{R}^d$, and is average-pooled from all token-level vectors of $s_i$. 

Second, we concatenate all sentence-level vectors \{$h_i$\}. Then a one-hot vector $m^t$ of length $n$, indicating which sentence is critical, is derived by:
\begin{align}
m^t &\sim \textrm{Gumbel}(W[h_1, h_2, \ldots, h_n])
\label{eq:gumbel}
\end{align}
where $W\in\mathcal{R}^{n \times nd}$, ``Gumbel'' is Gumbel-Softmax \cite{maddison2016concrete}, calculating a Gumbel distribution over the linear model predictions and samples categorical one-hot value from it. We use Gumbel-Softmax because it enables gradient backpropagation to help train the system end-to-end.

Since $m^t$ is $n$-dimensional one-hot vector; it only picks a single critical sentence.
To aggregate more potentially useful information from $I$, we do the Gumbel sampling procedure $k$ times (where set $k$ as 2 in our experiments) and take the element-wise union of \{$m^t$\}, $t=[1, \cdots, k]$.
Accordingly, the final mask $m$ is a \(k^{'}\)-hot vector (\(k^{'} \leq k\)) with each $m_i$ as: 
\begin{align}
m_i = \cup_{t=1}^k m^t_i
\label{eq:union}
\end{align}
Therefore, $m$ enables the model to pick at most $k$ critical sentences in $I$. As shown in Figure~\ref{fig:overall architecture}, $I=\{s_1, s_2, s_3, s_4\}$, and $\{s_1, s_3\}$ are critical sentences.

\begin{table*}[t!]
 \setlength{\belowcaptionskip}{-10pt}
 \setlength{\abovecaptionskip}{5pt}
\centering
\resizebox{0.81\linewidth}{!}{
\begin{tabular}{lll|c|c|c}

\toprule

          &&        & \exactmatch                    & \rougeL                & \rougeL (overall)      \\ 

\midrule

\multicolumn{3}{l|}{{\color[HTML]{656565}GPT-4 \cite{OpenAI2023GPT4TR}  } }            & {\color[HTML]{656565}64.51\tiny{(\(\pm\)2.56)} }          & {\color[HTML]{656565}59.36\tiny{(\(\pm\)2.24)} }         & {\color[HTML]{656565}62.96\tiny{(\(\pm\)2.08)} }         \\

\multicolumn{3}{l|}{{\color[HTML]{656565}ChatGPT \cite{openai2022chatgpt}  }  }           & {\color[HTML]{656565}46.90\tiny{(\(\pm\)2.23)} }          & {\color[HTML]{656565}56.82\tiny{(\(\pm\)3.10)} }         & {\color[HTML]{656565}52.41\tiny{(\(\pm\)2.30)} }         \\

\midrule

\multicolumn{3}{l|}{SeqGAN \cite{yu2017seqgan}  }             & 24.50\tiny{(\(\pm\)1.13)}           & 31.19\tiny{(\(\pm\)2.09)}          & 27.55\tiny{(\(\pm\)1.32)}          \\
\multicolumn{3}{l|}{ReCross \cite{lin2022unsupervised}   }       & 28.95\tiny{(\(\pm\)0.45)}          & 38.81\tiny{(\(\pm\)0.92)}           & 33.88\tiny{(\(\pm\)0.58)}          \\
\multicolumn{3}{l|}{MetaICL (SeqGAN) \cite{min-etal-2022-metaicl}   }     & 24.28\tiny{(\(\pm\)0.98)}          & 33.65\tiny{(\(\pm\)1.87)}          & 28.14\tiny{(\(\pm\)1.22)}         \\
\multicolumn{3}{l|}{MetaICL (ReCross) \cite{min-etal-2022-metaicl} }   & 14.98\tiny{(\(\pm\)0.42)}          & 21.63\tiny{(\(\pm\)0.83)}          & 20.74\tiny{(\(\pm\)0.40)}          \\

\multicolumn{3}{l|}{\tkinstruct~\cite{wang2022benchmarking} }  & 28.56\tiny{(\(\pm\)0.39)}          & 39.35\tiny{(\(\pm\)0.85)}          & 33.64\tiny{(\(\pm\)0.47)}          \\

\midrule

\multirow{5}{*}{\rotatebox{90}{\begin{tabular}{c} \methodname\end{tabular}}}& \multicolumn{2}{l|}{Strategy I}   & 29.67\tiny{(\(\pm\)0.43)}                   & 39.54\tiny{(\(\pm\)0.90)}                   & 34.98\tiny{(\(\pm\)0.57)}                   \\
& \hspace{1em} \multirow{4}{*}{w/ Strategy II} &  \textit{ranking ori}  & 29.98\tiny{(\(\pm\)0.87)}                   & 41.79\tiny{(\(\pm\)1.08)}                   & 35.62\tiny{(\(\pm\)0.76)}                   \\
&& \textit{ranking del}  & 28.68\tiny{(\(\pm\)1.04)}                   & 41.86\tiny{(\(\pm\)1.21)}                   & 34.46\tiny{(\(\pm\)0.89)}                   \\
&& \textit{ranking null} & 29.34\tiny{(\(\pm\)0.92)}                   & 42.13\tiny{(\(\pm\)1.13)}                   & 35.10\tiny{(\(\pm\)0.93)}                    \\
&&  \textit{ranking all}  & \textbf{30.58}\tiny{(\(\pm\)0.83)} & \textbf{43.55}\tiny{(\(\pm\)1.02)} & \textbf{36.70}\tiny{(\(\pm\)1.14)} \\ 

\bottomrule

\end{tabular}
}
\caption{Main results. Numbers of different methods were calculated from three random runs. We also put LLMs' performances (GPT-4, etc.) here for reference (i.e., upper bound). Please see the appendix for the baselines' details.}

\label{tab:main results}
\end{table*}

\paragraph{Strategy II: ranking-based objective.} In a conventional text-to-text generation, we mainly optimize the probability, through negative log-likelihood ($\mathcal{L}_{nll}$), of generating the gold output. In zero-shot instruction following, when we are aware of which sentences in the $I$ are crucial, in addition to applying the standard loss $\mathcal{L}_{nll}$, we can further take a ranking loss to make sure more informative instructions ($I^+$) lead to gold outputs with higher probabilities than less informative ones ($I^-$).\footnote{The motivation is that, given the informative $I^+$, the models can still ignore the beneficial parts selected by Strategy I~\citep[cf.][]{mishra2022cross}. Thus, Strategy II further forces the models to pay attention to those crucial parts (textual differences between $I^+$ and $I^-$) by producing different probabilities.} Specifically, we can build ($I^+$, $I^-$) pairs in three ways:

\textbullet\enspace \textbf{Repeat vs. Origin} (origin): $I^+$ is [$s_1$, $s_2$, $s_3$, $s_4$, [REP], $s_1$, $s_3$, [REP]]. This means $\{s_1, s_3\}$ will be repeated in the input instruction, and the special token [REP] can help tell the model which part is highlighted. $I^-$ is [$s_1$, $s_2$, $s_3$, $s_4$];

\textbullet\enspace \textbf{Repeat vs. Delete} (delete): $I^+$ is [$s_1$, $s_2$, $s_3$, $s_4$, [REP], $s_1$, $s_3$, [REP]], $I^-$ is $I$ when those critical sentences are masked, i.e., [$s_2$, $s_4$];

\textbullet\enspace \textbf{Repeat vs. Null} (null): $I^+$ is [$s_1$, $s_2$, $s_3$, $s_4$, [REP], $s_1$, $s_3$, [REP]], and $I^-$ is an empty string.

Let's use $f_{I^+}(y|x)$ and $f_{I^-}(y|x)$ to denote the probabilities of generating the gold output $y$ given the input $x$ and the instruction. Then our ranking loss $\mathcal{L}_{rank}$ is implemented as:
\begin{equation}
    \mathcal{L}_{rank} = \mathrm{max}(0, \alpha-f_{I^+}(y|x)+f_{I^-}(y|x))
    \label{eq:ranking_loss}
\end{equation}
where $\alpha$ controls the probability margin, and $f_*(y|x)$ is the average of word-level probabilities on the decoder side. The final loss of our model \methodname~is $\mathcal{L}=\mathcal{L}_{nll}+\beta\cdot\mathcal{L}_{rank}$. Different approaches to generating ($I^+$, $I^-$) pairs can specify the $\mathcal{L}_{rank}$ as: $\mathcal{L}_{origin}$, $\mathcal{L}_{delete}$, or $\mathcal{L}_{null}$~(as shown in Figure~\ref{fig:overall architecture}). We will study their individual and joint contributions in experiments. When testing, we generate the final prediction on ``\textit{Repeat}''.

\section{Experiments}\label{sec:exp}

\paragraph{Dataset.} We work on the benchmark \dataname~\cite{wang2022benchmarking}, which contains 1,040 diverse English tasks (921 in $train$ and 119 unseen tasks in $test$). We follow~\citet{wang2022benchmarking} only using 756 tasks in $train$ to train the final model. Each task is expressed by an instruction, originally consisting of a paragraph-level task definition and a couple of positive\&negative examples, and a large set of input-output instances. To satisfy our setting, we only use definitions as instruction~$I$. The average definition length is 65.73 by words (4.09 by sentences). Those classification and generation tasks are respectively evaluated by \exactmatch~and \rougeL~\cite{lin2004rouge}. We also report \rougeL~(overall), which calculates the \rougeL~on both classification and generation tasks, to reflect an overall estimation. More dataset and metric details can be found in Appendix and Table~\ref{tab:data statistic}.


\paragraph{Baselines.} \label{subsec:baselines} Since prior systems for few-shot instruction following need examples in instructions, in order to apply them to a zero-shot setting, we try to generate silver examples for them. For this thread, our baselines include  (i) \textit{SeqGAN} \cite{yu2017seqgan}: Using GAN to generate silver $y$ by utilizing task definition and $x$; (ii) \textit{ReCross} \cite{lin2022unsupervised}: Retrieving similar examples from the training set using task definition and $x$; (iii) \textit{MetaICL} \cite{min-etal-2022-metaicl}: Meta-learning given task definition and a few examples. Due to the different resources of examples, \textit{MetaICL} is specified to \textit{MetaICL} (SeqGAN) and \textit{MetaICL} (ReCross). Another baseline concatenates task definition, examples, and $x$ in the encoder to decode $y$, namely the prior state-of-the-art system T$\mathbf{k}$-I{{\small{NSTRUCT}} \cite{wang2022benchmarking}}. More details about baselines are in the Appendix.

\paragraph{Our model implementation.} We follow \citet{wang2022benchmarking} using T5-base~\cite{raffel2020exploring} for all experiments. Please refer to Appendix and Table~\ref{tab:Hyperparameters} for more experimental settings (e.g., hyper-parameters and computational cost). 

\paragraph{Results.} Table \ref{tab:main results} summarizes the results on zero-shot instruction following.
Overall, our approach shows successive performance improvements by adding the two proposed strategies and gains state-of-the-art results by adopting them jointly, proving the effectiveness of our method.
%
Worth noting that the {T$\mathbf{k}$-I{{\small{NSTRUCT}}}} can be regarded as our backbone, and after adding strategy I, our method has already improved by 1.34 \rougeL~(overall) score, indicating the benefits of highlighting crucial sentences. Moreover, we gain further performance improvements by adding strategy II, because the ranking objective trains the model to discriminate the differences in the inputs, thus it drives the model to understand the highlighted information rather than simply ignoring them~\cite{webson-pavlick-2022-prompt}.
As for MetaICL, due to the huge task differences between $train$ and $test$ (as shown in Table \ref{tab:data statistic}), those silver examples generated or retrieved by using the $train$ do not provide the in-distribution patterns~\cite{min2022rethinking},\footnote{We also observed the low instance similarities predicted by ReCross between $train$ and $test$.} leading to sub-optimal or even worse performances, cf. MetaICL (ReCross) vs. ReCross.
Note that, ReCross directly retrains the model with the retrieved examples and obtains relatively better results, however, it is still suffering from the drawbacks of few-shot instruction following in such a strict cross-task setting, so as SeqGAN.

\begin{table}[t]
 \setlength{\belowcaptionskip}{-10pt}
 \setlength{\abovecaptionskip}{5pt}
\centering
\small
\begin{tabular}{L{0.98\linewidth}}
\toprule
$\mathbf{I}$: \textcolor{MyBlue}{The answer will be ``yes'' if the provided sentence contains an explicit mention that answers the given question. Otherwise, the answer should be ``no''.} Instances where the answer is implied from the sentence using   ``instinct'' or ``common sense'' [$\cdots$] should be labeled as ``no''.

$\mathbf{y}$: Yes.

\textbf{\tkinstruct} $\hat{\mathbf{y}}$: March

\textbf{\methodname} $\hat{\mathbf{y}}$: Yes\\\hline

$\mathbf{I}$: Given a text passage, come up with an appropriate title for it. [$\cdots$] \textcolor{MyBlue}{The title should be 1-5 words long.}

$\mathbf{y}$: Nobel Peace Prize

\textbf{\tkinstruct} $\hat{\mathbf{y}}$: The Nobel Peace Prize is one of the five Nobel Prizes created by the Swedish industrialist, inventor, and armaments manufacturer Alfred Nobel.

\textbf{\methodname} $\hat{\mathbf{y}}$: Nobel Peace Prize\\\hline

$\mathbf{I}$: In this task, you're given an ambiguous question (which can be answered in more than one way). \textcolor{MyBlue}{Your task is to write a question that clarifies the given question in such a way that the generated question has one unique answer}.

$\mathbf{y}$: When was the National World War II memorial officially established?

\textbf{\tkinstruct} $\hat{\mathbf{y}}$: 1830

\textbf{\methodname} $\hat{\mathbf{y}}$: When was the memorial built?\\
\bottomrule

\end{tabular}
\caption{Effect of Strategy I. $\hat{\mathbf{y}}$: system output. The detected crucial sentences are highlighted in \textcolor{MyBlue}{blue}.}
\label{tab:effect-I}
\end{table}

\paragraph{Analysis.} 
\label{subsec:analysis}
We try to clear up three concerns.

\textbf{$\mathcal{Q}_1$: Did the detected critical sentences really contribute to the generation of gold outputs?} To answer $\mathcal{Q}_1$, we checked some examples where our system improves over the strongest baseline \tkinstruct. As shown in Table~\ref{tab:effect-I}, our approach can generally point out those crucial task-relevant sentences that are hardly encoded by the \tkinstruct, such as \textit{output space} (the first example), \textit{length constraint} (the second example), and \textit{types of output} (the last example). With the help of such highlights, our system can produce outputs that are better aligned with the task definitions, while \tkinstruct~often violates the requirements of instructions.

\begin{table}[!t]
 \setlength{\belowcaptionskip}{-10pt}
 \setlength{\abovecaptionskip}{5pt}
\centering
\small
\begin{tabular}{L{0.98\linewidth}}

\toprule



$\mathbf{I}$: Generate an overlapping word between the given two sentences. [$\cdots$] \textcolor{MyBlue}{You must generate significant words which \textbf{are not} the stop words like ``the'' or ``of", etc.}  \newline $\mathbf{x}$: $s_1$: Amphibians have permeable skin that easily absorbs substances from the environment. $s_2$: Amphibians begin their lives in the water. \newline $\mathbf{y}$: Amphibians || $\hat{\mathbf{y}}$:the \newline \textcolor{red}{Error type: negation}\\\hline


$\mathbf{I}$: Two analogies that relate items to whether they are trash or treasure is given in the form ``A : B. C : ?" [$\cdots$] \textcolor{MyBlue}{``A : B'' relates item A to whether it is trash or treasure, as specified by B. }[$\cdots$] \newline $\mathbf{x}$: baby : treasure. leaf : ? \newline $\mathbf{y}$: trash || $\hat{\mathbf{y}}$: relates item A to whether it is trash or treasure \newline \textcolor{red}{Error type: pattern copy}\\\hline


 $\mathbf{I}$: [$\cdots$] If it is about requesting something, generate 'REQUEST'. [$\cdots$] \textcolor{MyBlue}{If it is about informing something, generate ``INFORM''.} \newline $\mathbf{x}$: Please tell me do you have any particular date for the event? \newline $\mathbf{y}$: REQUEST || $\hat{\mathbf{y}}$: INFORM \newline \textcolor{red}{Error type: incomplete critical sent. detection}\\ 

\bottomrule

\end{tabular}
\caption{The error patterns by our system.
We highlight the crucial sentences in the instructions with \textcolor{MyBlue}{blue}, and mark the error type as \textcolor{red}{red}. }
\label{tab:error}
\end{table}

\textbf{$\mathcal{Q}_2$: Could ranking objective really improve the probability of gold outputs?} Regarding $\mathcal{Q}_2$, we test our model on all \textsc{Test Tasks} with two versions of task instructions: repeat vs. origin.  For each version, we calculate the corresponding probability of the ground truth output by averaging token-level probabilities in the output string. Our model can produce a higher ground-truth probability once ``repeat'' instruction is adopted (score: 0.59) than the ``origin'' definition (score: 0.11),\footnote{Average from three random seeds experiments.} demonstrating the effectiveness of our Strategy II.

\textbf{$\mathcal{Q}_3$: Error patterns of our systems.} We randomly pick up 200 instances from the $test$ and summarize three main error patterns of \methodname, as shown in Table~\ref{tab:error}. (i) \textit{Negation}. As the first example in Table~\ref{tab:error} shows, even though the model is able to detect the sentence that has a specific requirement ``generate significant words which are not the stop words $\cdots$'', the negation ``are not'' was not successfully comprehended by the system. Unfortunately, negation understanding has increasingly been a challenge in NLP \cite{DBLP03871,yin2022contintin,khashabi2022reframing}. (ii) \textit{Pattern copy}. The second example shows the system sometimes copies a span from the definition, especially when the definition string, e.g., ```A : B' relates item A to whether it is trash or treasure, as specified by B.'', matches the format of $x$, e.g., ``baby : treasure. leaf: ?''. This resembles demonstration-driven in-context learning, where researchers found pattern match is a key factor of success~\cite{min2022rethinking}. (iii) \textit{Incomplete critical sentence detection}. It is possible that our system detects partial sentences that are critical. As a result, the system is biased toward the requirement of highlighted sentences. Rather than using a hard masking scheme, our future work will explore a soft-masking technique so that no instruction parts will be clearly ignored.

\section{Conclusion}

In this paper, we focused on zero-shot instruction following, where we only adopted the task definitions as the instructions to help the model perform cross-task generalization. Expressly, our method pointed the critical sentences out of the lengthy definitions and highlighted them explicitly. In addition, we further designed a ranking objective to improve the instruction grasp of the LMs. We also conducted thorough analyses to help future research on zero-shot instruction following.



\bibliography{custom}
\bibliographystyle{acl_natbib}

\appendix
\newpage
\clearpage

\begin{table}[!t]
\centering
\resizebox{\linewidth}{!}{
\begin{tabular}{l|c}
\hline
\textbf{Hyper-parameters}  & \textbf{Range}           \\ \hline
lr for T5 & [5e-6, 1e-5, \textbf{5e-5}, 1e-4]   \\ \hline
lr for Pointer Networks & [5e-5, 1e-4, \textbf{3e-4}, 5e-4]   \\ \hline
lr for Encoder & [1e-6, \textbf{5e-6}, 1e-5, 5e-5]   \\ \hline
\(\alpha_{origin}\)  & [0.001, 0.003, \textbf{0.01}, 0.03, 0.1] \\ \hline
\(\alpha_{delete}\)  & [0.001, 0.003, 0.01, \textbf{0.03}, 0.1] \\ \hline
\(\alpha_{null}\)  & [0.01, 0.03, \textbf{0.1}, 0.3] \\ \hline
\(\beta\) & [0.01, 0.05, 0.1, 0.5, \textbf{1}] \\ \hline
\(k\) & {[}1, \textbf{2}, 3, 4, 5{]} \\ \hline
Pooling Function & {[} \textbf{average}, max {]} \\ \hline
\end{tabular}
}
\caption{The hyper-parameters trialed in tuning our models. The best ones adopted in our final experiments are highlighted in boldface. Here, ``lr'' denotes the learning rate; \(\alpha\) is the probability margin in equation~\ref{eq:ranking_loss}, there are three different \(\alpha\) according to three ranking losses; \(\beta\) is a coefficient that controls the
influence of the ranking losses; and \(k\) is the sampling times in equation~\ref{eq:union}.}

\label{tab:Hyperparameters}
\end{table}
%

\section*{Appendix A. Expanded Technique Details}

Due to the length limitation, we have to elaborate on some other important details of our approach in this section, including four different instructions in Figure~\ref{fig:overall architecture} and how we enable end-to-end optimization. As we have illustrated in Figure \ref{fig:overall architecture}, our approach consists of two parts, corresponding to \textbf{Strategy I} and \textbf{Strategy II} in Section \ref{sec:method}, respectively.

\textbf{Strategy I} (the left dashed box in Figure \ref{fig:overall architecture}) first encodes and converts all the sentences in a definition to sentence-level representations. Then, we adopt pointer networks followed by a Gumbel-Softmax layer to predict a binary vector for these representations, where ``1'' means the corresponding sentence contains crucial task-relevant information and should be attended by the LMs. In order to pick up more potentially useful information, we repeat the Gumbel sampling several times and take the element-wise union of the sampling results as the final decision of strategy I. It is worth noting that the encoder of this phase shares the same model structure as the encoder of the LMs to keep similar internal features of the downstream procedure~\cite{lin2022unsupervised}. However, they are optimized individually.

\textbf{Strategy II} (the right solid box in Figure \ref{fig:overall architecture}) regards the output binary vector of strategy I as a sentence-level mask matrix and constructs four different instructions accordingly: (1). \textbf{\textit{Repeat}} indicates the definition in which the critical parts are repeated and highlighted. Practically, we repeat the whole definition once (surrounded by a special token ``[REP]'') and use the binary vector from the strategy I as the attention mask matrix in the Transformers~\cite{vaswani2017attention}; (2). \textbf{\textit{Origin}} is the original definition without any modifications; (3). \textbf{\textit{Delete}} denotes the definition where the critical parts are masked. Similar to \textit{Repeat}, we actually encode the whole definition and use the invert of the binary vector to mask the critical information; (4). \textbf{\textit{Null}} means that there are no instructions provided. Intuitively, if the model can truly understand the prefixed instructions, it shall discriminate these text differences and produce better results on the inputs with informative instructions (i.e., \textit{Repeat}) than the others (i.e., \textit{Origin}, \textit{Delete}, and \textit{Null}).\footnote{Unlike the \textit{Repeat}, we do not use any special tokens in the other instructions (``[DEL]'', ``[NULL]'', etc.) to avoid introducing shortcuts to the model~\cite{du2021towards}.} Therefore, besides the standard negative log-likelihood \(\mathcal{L}_{nll}\), there are three additional ranking losses in total, namely \(\mathcal{L}_{origin}\), \(\mathcal{L}_{delete}\), and \(\mathcal{L}_{null}\).

Notably, our system can be optimized end-to-end because we incorporate the decision of strategy I by utilizing the attention mask mechanism in the LMs of strategy II.

\section*{Appendix B. Experimental Details}

For hyper-parameters, we use segmented learning rate (5e-5 for T5, 3e-4 and 5e-6 for the pointer networks and encoder, respectively) optimized with Adam \cite{kingma2014adam}.
As for the margins of ranking losses, we follow previous works employing structured margins to obtain a better representation space in LMs~\cite{wang2019ranked,9413437}.
Following~\citet{wang2022benchmarking}, after two epochs training on $train$, we evaluate our model on $test$ with the beam size equal to 1 (greedy decoding).
We present our hyper-parameters selection in Table \ref{tab:Hyperparameters}.
All the ranges of these hyper-parameters are decided empirically, and we search for the best combination greedily by observing the \rougeL~score on the development set.
We use Hugging Face T5-base for all the experiments~\footnote{\url{https://huggingface.co/t5-base}} and utilize Spacy for sentence segmentation.\footnote{\url{https://github.com/explosion/spacy-models/releases/tag/en_core_web_sm-3.4.1}}
It is notable that the definition length can be diverse, and it will extremely increase the computational burden if we let the pointer networks consider all the sentences in a definition. According to Table \ref{tab:data statistic}, we randomly select 5 sentences from the definition of each task as the candidates.

All of our code is implemented by using Python 3.8.0 and PyTorch 1.12.1~\footnote{\url{https://pypi.org/project/torch/}} with CUDA 11.6, and we utilize Hugging Face Transformers 4.18.0~\footnote{\url{https://github.com/huggingface/transformers/releases}} to train and evaluate our models.
%
%
We conduct all our experiments on Ubuntu 18.04 LTS using Intel(R) Core(TM) i9-10900KF CPU with 32 GB of memory, and employing NVIDIA RTX A5000 GPU with 24 GB of memory.
On the whole, there are about 332 million parameters in our models.
It takes about 12 hours to train and evaluate our models (2 epochs with batch size equal to 1). At the same time, the peak of GPU usage is 23GB.

\begin{table}[!t]
\centering
\Large
\resizebox{\linewidth}{!}{
\begin{tabular}{l|r|r|r}
\toprule

& \multicolumn{1}{c|}{\textbf{Train}} & \multicolumn{1}{c|}{\textbf{Dev}} & \multicolumn{1}{c}{\textbf{Test}} \\ 

\midrule

\#   of tasks          & 756            & 100                               & 119                               \\
\#   of instances      & 75,317         & 9,958                            &  11,810                            \\
\#   of task types     & 60             & 23                                & 12                                \\
\#   of domain types   & 101            & 24                                & 35                                \\
\#   of sources        & 243            & 46                                & 75                                \\
sources overlap with test set & 0.0\%   & 80.4\%                              & /                                 \\

\midrule

avg   def. length (words per task)     & 66.41 & 65.58 & 61.55 \\
avg   def. length (sentences per task) & 4.11 & 4.12 & 3.92 \\

\bottomrule

\end{tabular}
}
\caption{The dataset statistics.}
\label{tab:data statistic}
\end{table}

\section*{Appendix C. Dataset and Metrics}
\label{ap:data}

We show the statistics of the benchmark dataset in Table \ref{tab:data statistic}. 
We only focus on the English tasks and use the same data split policy as previous work \cite{wang2022benchmarking}, where all those tasks coming from the same sources as the test set are excluded from the training set (as shown in Table \ref{tab:data statistic}).
However, because no official development set is provided, we randomly select 100 tasks from those excluded tasks with a maximum of 100 instances per task, as the development set used in our experiments.
Similarly, we follow \citet{wang2022benchmarking} to use the first 100 instances per testing task and randomly choose 100 instances per training task.

As for the evaluation metrics, we follow \citet{wang2022benchmarking} utilizing \rougeL~\cite{lin2004rouge} and \exactmatch~\cite{rajpurkar2016squad} to evaluate the cross-task generalization performance of the text-to-text LMs.
To be specific, the \rougeL~reflects the string overlap between the answers and the predictions, while \exactmatch~measures the ratio of the number of correctly predicted examples.
Both of these metrics are widely adopted by previous works~\cite{2016arXiv160605250R,poria2021recognizing,gu2023page}. 
Since the \exactmatch~calculates the ratio of how many ground truth labels are generated, it is similar to the accuracy score.
%
Thus, we report the \exactmatch~score for those classification tasks in Table \ref{tab:main results}.
What's more, we use the same evaluation script as \citet{wang2022benchmarking} to compute these metrics.\footnote{\url{https://github.com/yizhongw/Tk-Instruct/blob/main/src/compute_metrics.py}}


\begin{table}[!t]
 \setlength{\belowcaptionskip}{-10pt}
 \setlength{\abovecaptionskip}{5pt}
\centering
\small
\begin{tabular}{L{0.98\linewidth}}
\toprule

$\mathbf{I}$: You are given two sentences and have to find if there is entailment or agreement of the Hypothesis by the Premise. [$\cdots$] \textcolor{MyBlue}{Your task is to return ``entails'' if the premise supports hypothesis else return ``neutral''.}

$\mathbf{y}$: entails

\textbf{\tkinstruct} $\hat{\mathbf{y}}$: calorie

\textbf{\methodname} $\hat{\mathbf{y}}$: entails \\ \hline


$\mathbf{I}$: Generate an appropriate title for the given text.~\textcolor{MyBlue}{The generated title must be short and include the main topic of the text.}~\textcolor{MyBlue}{The preferred titles are under fifteen words.}

$\mathbf{y}$: Case Logic Laptop roller bag

\textbf{\tkinstruct} $\hat{\mathbf{y}}$: This bag is great for carrying laptop, HP Printer, portable scanner, cables and supplies

\textbf{\methodname} $\hat{\mathbf{y}}$: bag for laptop \\ \hline


$\mathbf{I}$: In this task, you are given two questions about a domain.~\textcolor{MyBlue}{Your task is to combine the main subjects of the questions to write a new, natural-sounding question.}~For example, [$\cdots$].

$\mathbf{y}$: Did this president go to college in the state he was born in?

\textbf{\tkinstruct} $\hat{\mathbf{y}}$: this president

\textbf{\methodname} $\hat{\mathbf{y}}$: this president was born on the east coast? \\ \hline


$\mathbf{I}$: \textcolor{MyBlue}{Given a document, generate a short title of the document.}~\textcolor{MyBlue}{The title should convey the main idea/event/topic about which the document is being written.}~Note that URLs in the text have been replaced with [Link].

$\mathbf{y}$: Dutch politician on trial on hate speech charges

\textbf{\tkinstruct} $\hat{\mathbf{y}}$: Geert Wilders

\textbf{\methodname} $\hat{\mathbf{y}}$: Geert Wilders is on trial for hate speech \\

\bottomrule

\end{tabular}
\caption{More cases. The crucial sentences are in \textcolor{MyBlue}{blue}.}
\label{tab:more_cases}
\end{table}

\section*{Appendix D. Baselines}

As mentioned in Section~\ref{subsec:baselines}, we implement four baselines for a comprehensive comparison. 
As follows, we provide detailed implementation information.
Worth noting that we tune all the hyper-parameters of the baselines on the development set or use the default settings reported by the original paper.

\paragraph{SeqGAN} It regards the generation as a sequential decision procedure and uses the Reinforcement Learning (RL) rewards of an additional classifier to optimize the generator. 
The original SeqGAN is based on LSTM \cite{10.1162/neco.1997.9.8.1735}. 
In order to fair compare with the other models, we change the backbone to T5-base.
For training the SeqGAN, including the generator and classifier, we use the following steps:
(1). Pre-training: we first pre-train the T5-base on the benchmark dataset as the generator, that is, we concatenate the original definition with the task input (i.e., $x$) and drive the model to predict the output (i.e., $y$).
As for the classifier, we use Hugging Face bert-large-cased~\footnote{\url{https://huggingface.co/bert-large-cased}} to perform a sequence classification, namely predicting the binary label (i.e., ``0'' or ``1'') by encoding the task definition and the ($x$, $y$) pair produced by the generator;
(2). Adversarial training: We follow \citet{yu2017seqgan} training the generator and classifier alternately.
Specifically, when generating each token, we employ Monte Carlo (MC)  search to complete the whole sequence and use policy gradient \cite{sutton1999policy} to optimize the generator.
After 20 steps of training on the generator (batch size equals 4), we use the silver answers predicted by the generator as the negative examples to train the classifier.
After adversarial training the generator with 5 epochs, we then use it to predict the instances of the unseen tasks in the test set (i.e., $\hat{y}$).  
Meanwhile, these ($x$, $\hat{y}$) pairs can also serve as examples for in-context learning (see MetaICL for more details).
%

\paragraph{ReCross} This is a retrieve-based method that utilizes the unlabeled examples of an unseen task to retrieve similar labeled examples from the training set. These retrieved examples can be further used for retraining the model. Similarly, they can also be used for in-context learning (i.e., MetaICL).
We follow the official implementation of \citet{lin2022unsupervised}.\footnote{\url{https://inklab.usc.edu/ReCross/}}
However, there are several differences between the original algorithm and our usage:
(1). We use the concatenation of definition and task input as the query and index for a fair comparison. We also believe the task definition can provide valuable semantics for the retrieval procedure;
(2). Instead of using RoBERTa \cite{liu2019roberta}, we train a Hugging Face bert-base-cased model as the Reranker,\footnote{\url{https://huggingface.co/bert-base-cased}} which has relatively better results in our experiments;
(3). We use T5-base as the backend of ReCross.

\paragraph{MetaICL} Following \citet{min-etal-2022-metaicl} and \citet{wang2022benchmarking}, we use task definition and two positive examples as instructions to train and test the T5-base model. While the test set examples are those silver examples produced by SeqGAN and ReCross, namely MetaICL (SeqGAN) and MetaICL (ReCross). All the other hyper-parameters are the same as what we use in the \tkinstruct.

\paragraph{\tkinstruct}

We use the official code and hyper-parameters of \citet{wang2022benchmarking}.\footnote{\url{https://github.com/yizhongw/Tk-Instruct}}
The only difference is that we use T5-base instead of T5-3B reported in their paper, due to the limited computational resources.
It is also worth noting that the original T$\mathbf{k}$-I{{\small{NSTRUCT}} is trained with positive demonstrations as additional instructions; in this paper, we solely use the task definition as the instruction of T$\mathbf{k}$-I{{\small{NSTRUCT}} to ensure a fair comparison.

\paragraph{ChatGPT \& GPT-4} For LLMs' performances, we use the scores reported by \citet{Lou2023MUFFIN} in Table~\ref{tab:main results}, where they concatenate the task instruction with input as a whole query of APIs. Please refer to \citet{Lou2023MUFFIN} for more details. 



\section*{Appendix E. More Cases}
We display more intuitive cases in Table~\ref{tab:more_cases}.

\section*{Appendix F. Limitations}
In this section, we summarize several limitations and broader impacts of this paper.
(1) As mentioned in Section \ref{sec:exp}, one limitation of this paper is that our approach is still difficult to fully encode the crucial information in the definitions, even if they are well highlighted, such as the negation expresses. Potential solutions include adopting an additional weighting strategy on the decisions of the pointer networks~\cite{see2017get}, adding a soft fusion mechanism in the LMs \cite{gao2021improving,tian-etal-2022-improving-english}, or proposing an automatic instruction reframing technology~\cite{khashabi2022reframing}.
(2) Meanwhile, since the task definition is usually a paragraph consisting of several sentences, this paper mainly focuses on detecting crucial sentence-level information. However, in some cases, task-relevant information should be better represented in a word-level or span-level format, such as the \textit{output space}. Therefore, our strategy can be further improved by using a hybrid-level pointer to satisfy the diverse real-world scenarios.
(3) Another potential future investigation is to analyze how LMs utilize the highlighted information in the instructions through human intuition, such as visualizing the multi-head attention score distribution of the transformers~\cite{ma2021contributions,ma2021gradts}, or probing the conflict between the in-context instruction and model's parametric knowledge~\cite{xie2023adaptive}.
We leave them as our future work.

\end{document}